\documentclass[sigconf]{acmart}

\AtBeginDocument{%
  }
\usepackage{cleveref}
\usepackage{colortbl}
\usepackage{subcaption}
\usepackage{multirow}
\usepackage{booktabs}
\usepackage{multirow}
\usepackage{graphicx}
\usepackage{balance}
\usepackage{hyperref}
\usepackage{url} 
\hypersetup{
    colorlinks=true,  %
}
\copyrightyear{2025}
\acmYear{2025}
\setcopyright{acmlicensed}\acmConference[MMSports '25]{Proceedings of the 8th International ACM Workshop on Multimedia Content Analysis in Sports}{October 27--28, 2025}{Dublin, Ireland}
\acmBooktitle{Proceedings of the 8th International ACM Workshop on Multimedia Content Analysis in Sports (MMSports '25), October 27--28, 2025, Dublin, Ireland}
\acmISBN{979-8-4007-1835-9/2025/10}
\acmDOI{10.1145/3728423.3759406}

\begin{document}

\title{Hand Held Multi-Object Tracking Dataset in American Football}

\author{Rintaro Otsubo}
\affiliation{%
  \institution{Keio University}
  \city{Yokohama}
  \country{Japan}
}
\email{rintarootsubo@keio.jp}

\author{Kanta Sawafuji}
\affiliation{%
  \institution{Keio University}
  \city{Yokohama}
  \country{Japan}
}
\email{sawafuji_kanta@keio.jp}

\author{Hideo Saito}
\affiliation{%
  \institution{Keio University}
  \city{Yokohama}
  \country{Japan}
}
\email{hs@keio.jp}

\renewcommand{\shortauthors}{Rintaro Otsubo, Kanta Sawafuji \& Hideo Saito}

\begin{abstract}
Multi-Object Tracking (MOT) plays a critical role in analyzing player behavior from videos, enabling performance evaluation. 
Current MOT methods are often evaluated using publicly available datasets.
However, most of these focus on everyday scenarios such as pedestrian tracking or are tailored to specific sports, including soccer and basketball.
Despite the inherent challenges of tracking players in American football, such as frequent occlusion and physical contact, no standardized dataset has been publicly available, making fair comparisons between methods difficult.
To address this gap, we constructed the first dedicated detection and tracking dataset for the American football players and conducted a comparative evaluation of various detection and tracking methods.
Our results demonstrate that accurate detection and tracking can be achieved even in crowded scenarios.
Fine-tuning detection models improved performance over pre-trained models.
Furthermore, when these fine-tuned detectors and re-identification models were integrated into tracking systems, we observed notable improvements in tracking accuracy compared to existing approaches.
This work thus enables robust detection and tracking of American football players in challenging, high-density scenarios previously underserved by conventional methods.

Our data is available at: \url{https://rinost081.github.io/MOTAF_page/}

\end{abstract}


\begin{teaserfigure}
  \includegraphics[width=0.75\textwidth]{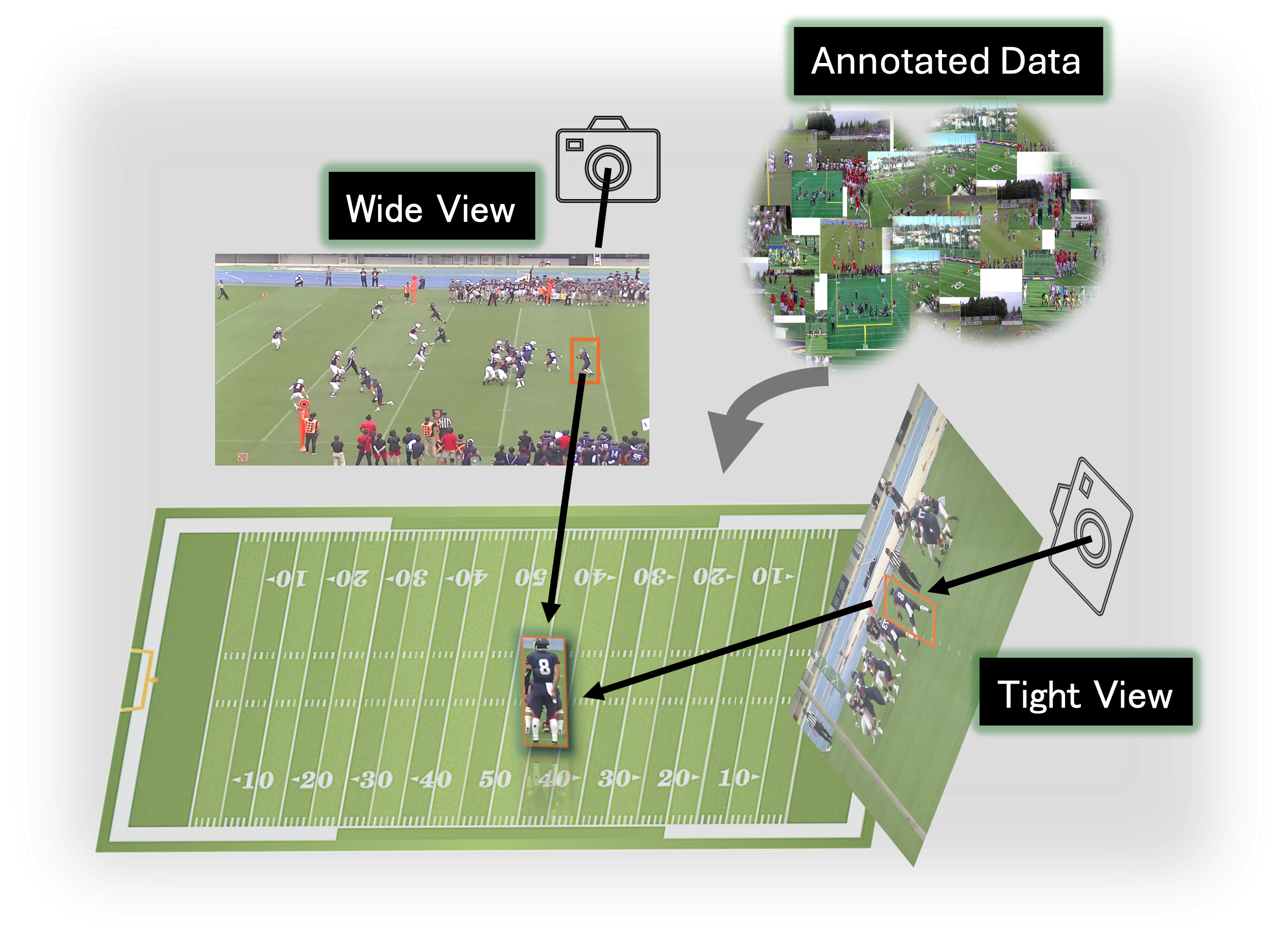}
  \centering
  \caption{Synchronized multi‐view captures from our American football dataset. The \emph{Wide View} and \emph{Tight View} illustrate two camera angles recording the same player in a blue uniform, highlighting precise spatial correspondence between views.}
  \label{fig:teaser}
  \Description{teaser}
\end{teaserfigure}


\maketitle

\section{Introduction}
American football presents a uniquely challenging environment for computer vision due to its crowded formations and fast player movements. 
At the professional level, the National Football League addresses these challenges by equipping players with sensors embedded in their uniforms, enabling precise real-time tracking throughout games.
However, our focus is on university-level American football, where such hardware-dependent systems are impractical due to its cost, setup complexity, and limited accessibility.
This highlights the importance of developing device-free, vision-based multi-object tracking (MOT) methods that can operate solely on video data, facilitating scalable and infrastructure-independent analysis across broader domains.
Device-free MOT in university-level American football offers a range of tangible applications, e.g., tracking each player's speed and distance for training, precise player segmentation for tactical video analysis, and detecting heavy impacts to alert medical staff immediately.

Numerous MOT datasets (MOTChallenge \cite{leal2015motchallenge} \cite{MOT16} \cite{MOTChallenge20} and KITTI \cite{geiger2012we} \cite{liao2022kitti} ) have been proposed, yet the majority focus on real-world scenarios such as pedestrian movement or urban driving scenes to evaluate tracking performance in surveillance and autonomous driving contexts.
In the sports domain, previous research works have primarily addressed soccer.
For instance, SSET \cite{feng2020sset} focuses on single-player tracking in soccer, while SoccerNet \cite{cioppa2022soccernet} extends to multi-player tracking in professional soccer games.
More recent works, such as SportsMOT \cite{cui2023sportsmot} and  TeamTrack \cite{scott2024teamtrack}, include multiple sports domains within a single dataset, aiming to evaluate the cross-domain generalizability of tracking models.
However, to the best of our knowledge, there is currently no publicly available dataset specifically designed for the detection and tracking of American football players.
Without a standardized benchmark for American football MOT, it is difficult to compare model performance across academic and industrial studies.
Moreover, domain-specific challenges remain underexplored, making it difficult to quantify their impact on tracking performance.

We propose the first American football dataset, where frequent occlusions and rapid player's movement are seen.
The dataset is built from live game footage collected from university-level matches and contains approximately 200,000 annotated bounding boxes across more than 10,000 frames.
Our dataset is taken from two views: Wide view and Tight view which are common angles to analyze videos in American football.
This dual-view setup enables research into viewpoint-aware tracking and robustness across camera angles.
We also include detailed dataset statistics to help characterize the dataset's structure and label distribution.
Using this dataset, we quantitatively and qualitatively evaluate the detection and tracking performance of several state-of-the-art MOT models.

In summary, the main contributions of this work are:
\begin{enumerate}
    \item We introduce the \textbf{first dataset} for Multi-Object Tracking (MOT) in \textbf{American football} from two views.
    \item Our dataset captures the unique visual challenges of American football—such as frequent occlusions and player collisions—offering a new benchmark for evaluating tracking robustness under dense and dynamic conditions.
    \item We apply existing MOT models to evaluate our dataset, and we investigate the influence of fine-tuning detection models and re-identification models.
\end{enumerate}

\begin{table}
\caption{Comparison of human MOT datasets. \\
         All datasets are single-domain, “–” indicates data not available.}
\centering
\small
\setlength{\tabcolsep}{5pt}
\resizebox{\linewidth}{!}{
\begin{tabular}{lccccl}
\toprule
\textbf{Dataset} & \textbf{Videos} & \textbf{Views} & \textbf{Bbox/Frame} & \textbf{Domain} \\
\midrule
MOT16 \cite{MOT16} & 14 & 1 & 26.1 & Pedestrians \\
MOT20 \cite{MOTChallenge20} & 8 & 1 & 123.2 & Pedestrians \\
DanceTrack \cite{sun2022dancetrack} & 100 & 1 & -- & Dance \\
SSET \cite{feng2020sset} & 80 & 1 & 1.00 & Soccer \\
SN-Tracking \cite{cioppa2022soccernet} & 201 & 1 & 16.18 & Soccer \\
\textbf{Ours} & 20 & \textbf{2} & 17.4 & \textbf{American Football} \\
\bottomrule
\end{tabular}
}

\end{table}

\section{Related Works}

\begin{figure*}
    \centering
    \includegraphics[width=\linewidth]{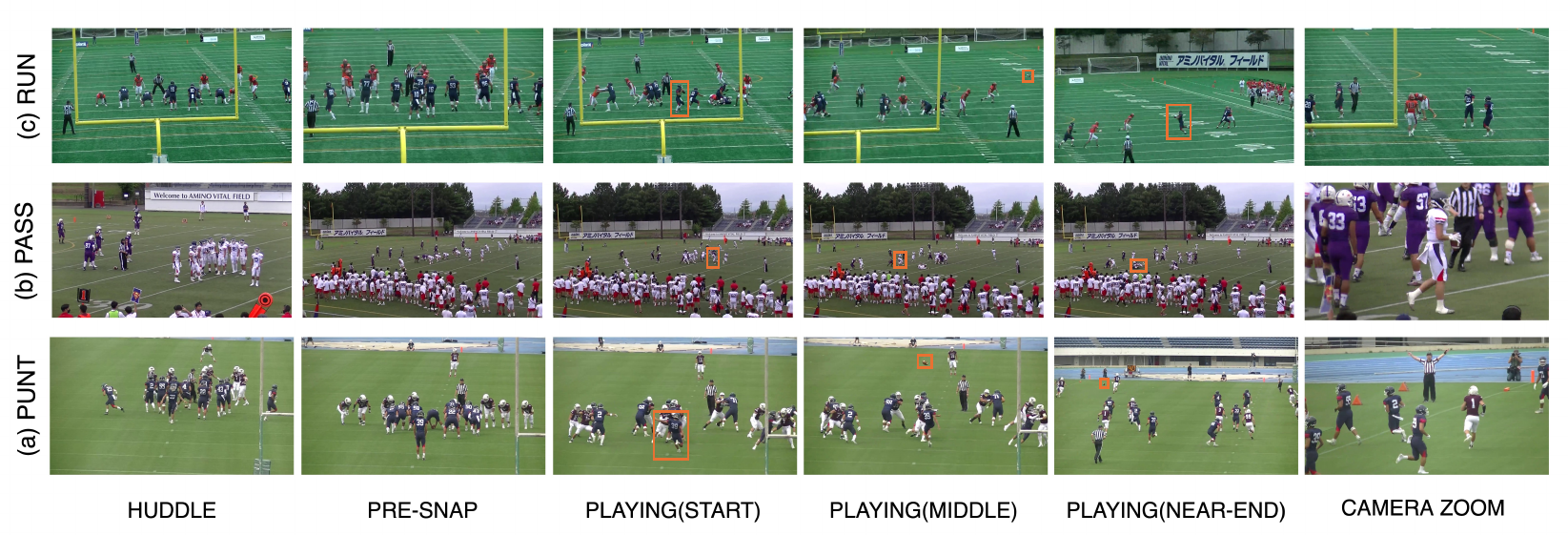}
    \caption{Examples of play categories by type (rows: pass, run, punt) and phase (columns: huddle to near-end). A ball or key players holding the ball are highlighted as an orange bounding box for clarity.}
    \label{play_kind}
    \Description{play kind}
\end{figure*}

\subsection{Detecting and Tracking Players}
Multi-Object Tracking (MOT) aims to track multiple objects from given videos.
A typical MOT system consists of two main components: an object detector, which identifies instances in each frame, and a data association algorithm, which links detections over time to form trajectories.
Early approaches to MOT relied on simple but effective matching algorithm.
One such method is Simple Online and Realtime Tracking (SORT) \cite{bewley2016simple}, which uses a Kalman filter to predict the location of bounding boxes in the next frame.
SORT has limitations in tracking performance in scenes with occlusions and non-linear motion.
To address this limitation, OCSORT \cite{cao2023observation} introduces an observation-centric association mechanism that improves robustness against occlusions and adapts better to irregular object motion.
Followed by, DeepOCSORT \cite{maggiolino2023deep}, which adds deep appearance features to OCSORT for more robust tracking, and StrongSORT \cite{du2023strong}, which employs deep appearance features.

Another popular method in MOT task is two-stage matching pipeline.
ByteTrack \cite{zhang2022bytetrack} proposed this two-stage matching pipeline to restore low confidence detections.
BoT-SORT (Bag of Tricks for SORT) \cite{aharon2022bot} is an extended and efficient method based on ByteTrack.
BoT-SORT addresses tracking challenges by integrating modules such as camera motion compensation (CMC) and re-identification (ReID) to improve tracking accuracy.
For association, BoT-SORT utilizes the minimum similarity between cosine distance or motion distance.
In contrast, ImprAsso \cite{stadler2023improved} uses both appearance similarity and motion information, using a weighted formulation controlled by a hyperparameter.

\subsection{Multi-Object Tracking Datasets}
Multi-Object Tracking datasets vary from a daily scene to specific sports scenes.
MOT15 \cite{leal2015motchallenge}introduces the first large publicly available dataset for Multi-Object Tracking (MOT) with a focus on pedestrian detection.
After that MOT17\cite{MOT16} and MOT20 \cite{MOTChallenge20} have been released, pushing MOT task forward by increasing the number of bounding boxes.
Other MOT dataset such as KITTI \cite{geiger2012we} and KITTI360 \cite{liao2022kitti} focus on autonomous driving tasks.

In sports scenes, SSET \cite{feng2020sset} creates a single player tracking dataset in soccer.
After that, SoccerNet-Tracking \cite{cioppa2022soccernet} proposes a large-scale MOT dataset for soccer.
SportsMOT \cite{cui2023sportsmot} and TeamTrack \cite{scott2024teamtrack} build MOT dataset for diverse sports scenes.
These dataset broadly analyze diverse sports scene including soccer, volleyball, and basketball, enabling the development and evaluation of robust general MOT models' ability under dynamic, occluded, and high speed conditions typical of real-world gameplay.

In this work, we propose an American football dataset for further sports analysis, where rapid directional changes, and frequent occlusions caused by protective gear and close contact introduce distinct tracking challenges.
All videos in our dataset are captured using handheld cameras, introducing motion blur that affects tracking accuracy.

\subsection{Person Re-Identification}
Person re-identification (ReID) refers to the task of matching the same individual across different camera views or video frames.
It is particularly challenging when individuals wear visually similar clothing, as is often the case in uniforms or team-based environments.
In the context of sports, this challenge becomes more severe in team sports where players wear identical uniforms.
Chong \textit{et al.} \cite{xiang2024person} introduces the Person in Uniforms Re-identification task, which focuses on distinguishing individuals wearing similar outfits.
In sports case, the SoccerNet community organizes a Player Re-Identification challenge \cite{Giancola_2022} where players wear same uniform, making it difficult to match the same players across different viewpoints.
Our scenario presents similar difficulties, since American football players on same team wear nearly identical uniforms, including helmets and protective padding, which significantly reduce visual cues for identity.
As a result, appearance-based person re-identification becomes particularly important for maintaining consistent player identities across frames.

\section{Dataset}
We created the first American football dataset for Multi-Object Tracking, where dynamic movements and collisions frequently occur.
This dataset consists of videos of an American football game captured by two hand-held cameras and annotated for a multi-object tracking task.

In this task, we aim to track players and referees on the American football field.
Therefore, although individuals such as audiences, substitute players, and photographers may appear in the footage, we focus exclusively on those present within the playing field.

\subsection{Data Collection}
We collected videos from live matches between K-University’s American football team and five opposing teams.
Each play was recorded from two distinct viewpoints (see \Cref{fig:teaser}) :
\begin{itemize}
\item 
Tight view: a close-up camera angle focused on the area near the ball. 
This view captures fine-grained actions occurring around the core of the play, however, tight view does not cover the entire field.

\item
Wide view: a zoomed-out angle that shows the entire field and includes nearly all players. This perspective is useful for understanding overall player formations and movements.
\end{itemize}

All matches in this dataset were recorded using hand-held cameras due to the organizer’s policy prohibiting the use of tripods. This setting simulates a casual recording environment, such as using a smartphone, rather than footage captured with pre-installed professional equipment. As such, the dataset offers an opportunity for amateurs to analyze the game in a realistic setting. We believe this can promote the broader adoption of advanced AI-based sports analysis technologies by the general public.

\subsection{Annotation Process}
We annotated bounding boxes around players and referees to detect them, and assigned unique IDs to each for tracking.
From the videos we took, we selected games against five different universities and picked two plays from each, resulting in 10 plays with both tight and wide views available for each.
We used the CVAT annotation tool to manage videos and generate bounding boxes.
In American football, each team consists of 11 players, and up to 4 referees may be present on the field.
Thus, we annotated a maximum of 26 bounding boxes per frame.

\subsection{Dataset Statistics and Analysis}
We manually annotated 11,411 frames, resulting in 188,711 bounding boxes for American football players and referees, and 505 unique tracks across all videos.

The dataset covers three main actions of American football plays -- run, pass, and punt -- as illustrated in \Cref{play_kind}.
Each play category presents unique visual characteristics.
During pass and punt plays, rapid camera transition make it difficult to consistently track American football players.
In contrast, run plays often feature players clustering around the ball, and dynamic tracklets can introduce significant occlusions, which challenge tracking algorithms.

\begin{figure}[t]
    \centering
    \includegraphics[width=0.6\linewidth]{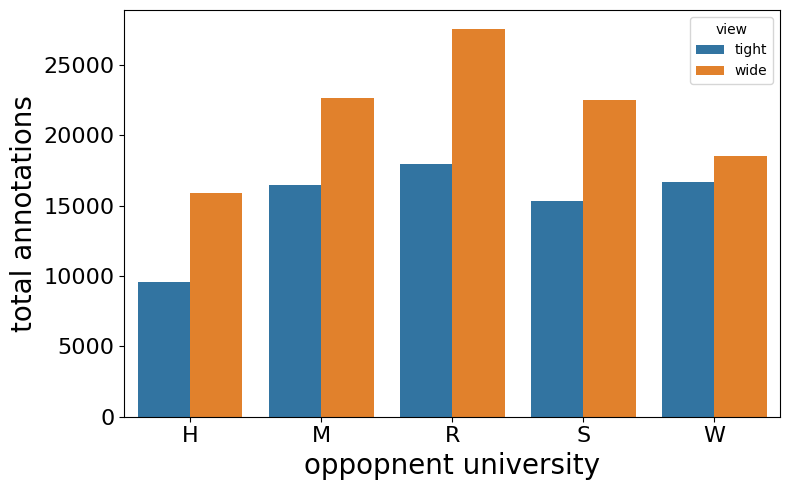}
    \caption{Total number of annotations per opponent university. Orange bars represent wide, and blue bars represent tight.}
    \label{fig:total_annotations}
    \Description{total_annotations}
\end{figure}

\noindent
\textbf{Detection}: 
As shown in \Cref{fig:total_annotations}, the match against R-University includes the highest number of bounding boxes, with over 25,000 annotations for the wide view and around 17,500 for the tight view.

\begin{figure}[t]
    \centering
    \includegraphics[width=0.6\linewidth]{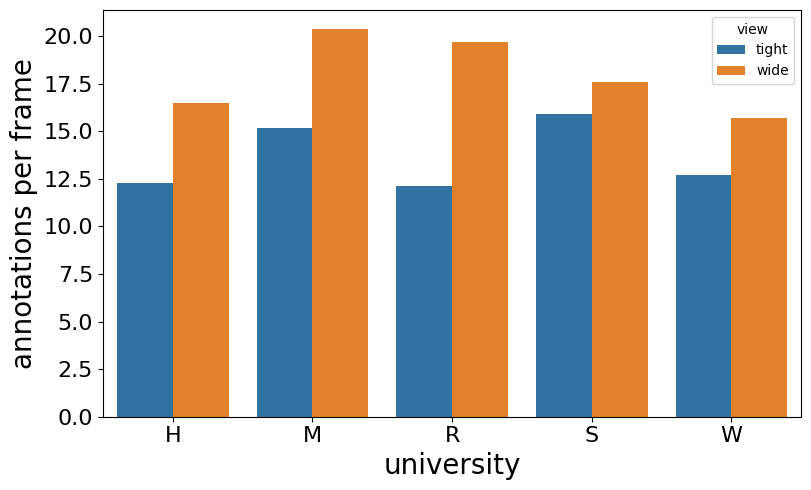}
    \caption{Average number of annotations per opponent university. Orange bars represent wide, and blue bars represent tight.}
    \label{fig:avg_annotations}
    \Description{avg_annotations}
\end{figure}

We further analyze spatial crowding across universities and camera angles by examining the average number of annotations per frame (\Cref{fig:avg_annotations})
Across wide view videos, the number of bounding boxes per frame ranges from 15 to 20, a difference of approximately 5.
In the tight view, the range is lower -- just over 15 at maximum and below 12.5 at minimum -- yielding a difference of roughly 2.5 annotations.
Overall, wide views consistently capture more players and referees, averaging 15.5 bounding boxes per frame, compared to 11.0 in tight views. 

Each play is divided into three temporal states; Huddle, Playing, and Camera Zoom.
These phases are illustrated in \Cref{play_kind}.
Huddle: players gather in close formation to discuss the upcoming play, this results in dense, highly occluded scenes.
Playing: the play is in motion, with players actively engaging.
Camera Zoom: after the play ends, the camera zooms in to identify the final ball handler, this phase included fewer visible participants.
\Cref{annotaions_per_state} shows the average annotated instances per frame across these phases.
For instance, M-University has notably high bounding boxes counts in both Huddle and Playing states, indicating dense formations.
Across all teams, the playing state consistently shows higher bbox counts per frame by 4.1 (Huddle) and 6.6 (Camera Zoom), representing more active players and referees on screen compared to Huddle or Camera Zoom states.

\begin{table}[t]
\caption{Average annotations in one frame per phase (huddle: players get together for instant discussion, playing: players actually play American football, camera zoom: we try to identify who involves plays at last) by opponent university}
\label{annotaions_per_state}
\centering
\begin{tabular}{lccc}
\hline
\textbf{university} & \textbf{huddle} & \textbf{playing} & \textbf{camera zoom} \\
\hline
H-University   & 10.7 & 15.4 & 14.0 \\
M-University   & 19.9 & 19.7 & 9.6  \\
R-University  & 14.5 & 18.1 & 9.7  \\
S-University  & 8.5  & 18.1 & 13.0 \\
W-University  & 13.0 & 15.8 & 9.0  \\
\hline
\rowcolor{gray!15}
Overall & 12.7 & 16.8 & 10.2 \\
\hline
\end{tabular}
\end{table}

\begin{figure}[t]
    \centering
    \includegraphics[width=0.95\linewidth]{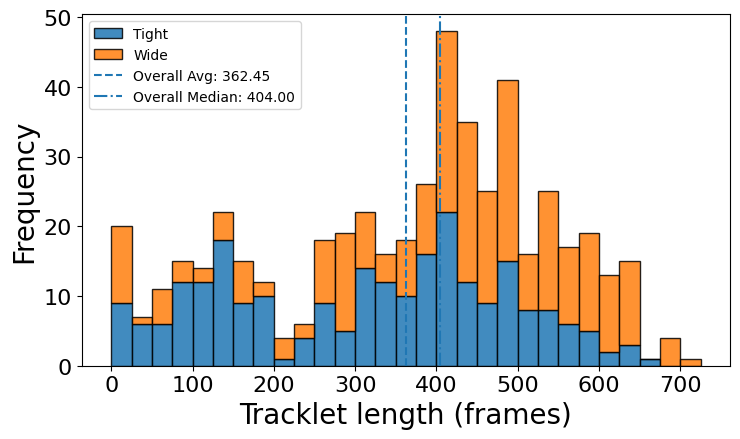}
    \caption{Tracklet length histogram per opponent university. Orange bars represent wide, and blue bars represent tight.}
    \label{fig:tracklet_length_histogram}
    \Description{tracklet_length_histogram}
\end{figure}

\noindent
\textbf{Tracking} :
\Cref{unique_ids} shows the number of unique IDs and average tracklet length across different universities, and \Cref{fig:tracklet_length_histogram}
illustrates the distribution of the tracklet lengths.

The matches involving H-University have the highest number of unique IDs at 113, while the minimum tracklet ids across all universities is 97, resulting in a maximum difference of only six tracklet IDs. 
This suggests that there are no substantial differences between the matches.

Regarding the tracklet lengths statistics, the mean tracklet length is smaller than the median, indicating a left-skewed distribution.
The distribution plot reflects this left-skewed tendency, implying that the dataset contains a greater number of short duration tracklets.

Among the universities, R-University has the longest average tracklet length, with a mean of 464.6 frames and a median of 480.8 frames, while H-University has the shortest, with a mean of 227.5 frames and a median of 277.2 frames.

\begin{table}[t]
    \caption{Tracking Statistics by University. Unique ids in each opponent games, and average tracklet length.}
    \label{unique_ids}
    \centering
    \label{tab:aggregated_tracking_statistics_univ}
    \begin{tabular}{l|c|cc}
        \toprule
        \textbf{University} & \textbf{Unique IDs} & \multicolumn{2}{c}{\textbf{Tracklet Length}} \\
        \cmidrule{3-4}
         & & \textbf{Mean} & \textbf{Median} \\
        \midrule
        H-University & 113 & 227.5 & 277.2 \\
        M-University& 97 & 401.3 & 435.2 \\
        R-University & 97 & 464.6 & 480.8 \\
        S-University & 97 & 387.1 & 430.6 \\
        W-University & 101 & 346.6 & 401.6 \\
        \hline
        \rowcolor{gray!15}
        {Overall} & {505} & {362.5} & {404.0} \\
        \hline
    \end{tabular}

\end{table}

\section{Experiments}
We conducted two main experiments to build baselines for detection and tracking tasks.

\subsection{Experimental Setup}
We train deep learning models on two NVIDIA TITAN
RTX GPUs, each with 24 GB of memory, utilizing CUDA 10.1. The
training system was optimized for performance using PyTorch

\noindent
\textbf{Dataset}: 
Our goal is to detect and track American football players, a setting characterized by frequent occlusions and rapid motions.
Although there are many existing MOT datasets, not only for daily-life but also for sports such as soccer and ice hockey, to the best of our knowledge, no MOT dataset exists for American football.
Therefore, we created the first MOT dataset for American football and conducted experiments using it.
We split it into train set, val set, and test set in a 5:3:2 ratio for model fine-tuning.

\subsection{Implementation Details}
\textbf{Detection}: 
In the Multi-Object Tracking setting, the detection-by-tracking approach has shown promising results.
Therefore, the detection model plays a crucial role in tracking American football players.
We fine-tuned YOLO and RTDETR(\textbf{R}eal-\textbf{T}ime \textbf{De}tection \textbf{Tr}ansformer) models for player detection.
Specifically, we fine-tuned YOLO models from version 8 to version 12, as well as RTDETR-x, and RTDETR-l, to identify the best-performing models for our dataset.

We fine-tuned each model for 50 epochs with a batch size of 16 and an input image resolution of 640×640.
The initial learning rate was set to 0.01, and it was scaled down by a factor of 0.01.
For evaluation, we used Recall, Precision, and mAP for the metrics.

\noindent
\textbf{Tracking}: 
For tracking performance, we compared Multi-Object Tracking models: BoostTrack~\cite{stanojevic2024boosttrack}, BoT-SORT, ByteTrack, DeepOCSort, Imprassoc, OCSORT, and StrongSORT for comparison.
These models incorporate both detection and re-identification components.
We fine-tuned the re-identification models to improve the performance of the MOT models.
The re-identification models were fine-tuned for 50 epochs with a batch size of 32.
We set the initial learning rate as 3e-4.
The evaluation metrics used are HOTA, MOTA, and IDF1, which respectively measure higher-order tracking accuracy, overall multi-object tracking performance, and identity preservation.

\begin{table}[t]
\centering
\caption{Detection performance on our test set. For each metric, values in parentheses indicate the absolute improvement compared to the same model trained without fine--tuning. Bold = best performance per metric.
}
\resizebox{\linewidth}{!}{
\begin{tabular}{l|c|c|c|c}
\hline
\textbf{Model} & \textbf{Recall} & \textbf{Precision} & \textbf{mAP50-95} & \textbf{mAP50} \\
\hline
YOLOv8   & 0.48 (+0.11) & {0.78 (+0.13)} & 0.48 (+0.11) & {0.80 (+0.23)} \\
YOLOv9   & 0.45 (+0.08) & 0.70 (+0.05) & 0.45 (+0.08) & 0.74 (+0.16) \\
YOLOv10  & {0.49 (+0.12)} & 0.76 (+0.07) & {0.49 (+0.12)} & 0.79 (+0.22) \\
YOLOv11  & 0.47 (+0.08) & 0.73 (+0.08) & 0.47 (+0.08) & 0.77 (+0.16) \\
YOLOv12  & 0.48 (+0.09) & 0.74 (+0.07) & 0.48 (+0.09) & 0.77 (+0.17) \\
RTDETR   & \textbf{0.50} (+0.13) & \textbf{0.82} (+0.10) & \textbf{0.50} (+0.13) & \textbf{0.84} (+0.25) \\ 
\hline
\end{tabular}
}
\label{quantitative_result}
\end{table}

\subsection{Detection} \label{ex_detection}
\textbf{Quantitative Results}:
We fine-tuned YOLO and RTDETR models on our training and validation sets, and then evaluated on the test set.
For comparison, we also evaluated the performance of the pretrained YOLO and RTDETR models on the same test set.

\Cref{quantitative_result} shows the performance of both pretrained and fine-tuned YOLO and RTDETR models on our test set.
Although we fine-tuned each YOLO model variant individually --including v8 \cite{yolov8_ultralytics} (n, s, m, l, x), v9 \cite{wang2024yolov9} (e, s, m, t), v10 \cite{wang2024yolov10} (s, n, m, l, x), 11 \cite{yolo11_ultralytics} (n, s, m, l, x),  v12 \cite{tian2025yolov12} (n, s, m, l, x), as well as RTDETR-l, and RTDETR-x models \cite{zhao2024detrs} -- the results in \Cref{quantitative_result} summarize their average performance under the representative names YOLOv8 through YOLOv12, and RTDETR. 
The values in parentheses indicate the absolute improvement compared to the corresponding model trained without fine-tuning.

Across all metrics and models, the fine-tuned versions demonstrate superior performance compared to their pretrained counterparts, indicating the effectiveness of domain-specific adaptation to our dataset.
Overall, RTDETR consistently outperforms all other models across every metric, suggesting it is the most effective detection model for our dataset.
YOLOv10 and YOLOv12 also achieve competitive results, although they fall slightly short of RTDETR’s scores.

Looking first at Recall, RTDETR achieves the highest score at 0.50,  followed closely by YOLOv10 (0.49) and YOLOv12 (0.48).
YOLOv9 records the lowest recall at 0.45. In terms of Precision, RTDETR again leads with 0.82, while YOLOv8 and YOLOv10 follow with 0.78 and 0.76 respectively.
As for mAP50-95, RTDETR attains the best score of 0.50, narrowly surpassing YOLOv10 (0.49) and YOLOv8 (0.48). A similar pattern emerges for mAP50, where RTDETR achieves 0.84 -- the highest among all --- compared to 0.80 by YOLOv8 and 0.79 by YOLOv10.
Furthermore, RTDETR exhibits the largest improvement over its pretrained version, particularly in mAP50 with an increase of +0.25, highlighting the strong benefit of fine-tuning.
In summary, while all models benefit from fine-tuning, RTDETR consistently delivers the strongest performance across all detection metrics, establishing itself as the most effective model among those evaluated.

\begin{figure*}
    \centering
    \includegraphics[width=\linewidth]{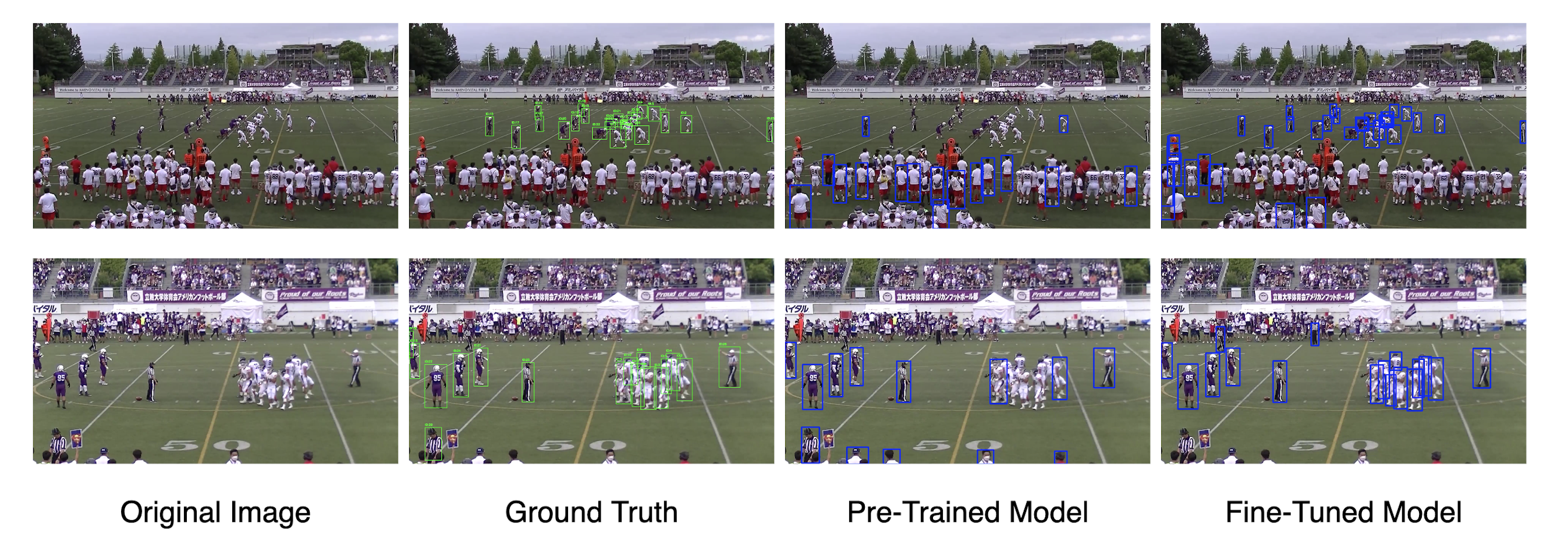}
    \caption{Qualitative Results on Detect players and referees}
    \label{fig:qualitative}
    \Description{qualitative}
\end{figure*}

\noindent
\textbf{Qualitative Results}:
\Cref{fig:qualitative} shows qualitative results on our test set, comparing the performance of the pre-trained and fine-tuned models.

As shown in the images, the pre-trained model often fails to detect players in crowded regions or under partial occlusions.
In contrast, the fine-tuned model successfully captures most players, including those partially occluded or tightly clustered.
This indicates that the fine-tuning significantly improves robustness in challenging scenarios, such as dense formations or overlapping players, which are common in American football.
Additionally, the bounding boxes from the fine-tuned model exhibit better alignment with ground-truth annotations, both in terms of location and size, leading to more accurate detection coverage.

\subsection{Tracking}
\begin{table}
\caption{Comparison of tracking performance across pre-trained MOT models and fine-tuned (FT) BoT-SORT. Parentheses indicate performance gain over the vanilla BoT-SORT. Bold = Best performance per metric, Underline = Second best performance per metric}
\label{tracking_result}
\centering
\begin{tabular}{l|c|c|c}
\hline
\textbf{Method} & \textbf{HOTA↑} & \textbf{MOTA↑} & \textbf{IDF1↑} \\
\hline
BoostTrack~\cite{stanojevic2024boosttrack}            & 25.3 & 22.0 & 29.1 \\
ByteTrack~\cite{zhang2022bytetrack}             & 28.7 & \underline{28.0} & \underline{34.4} \\
DeepOCSORT~\cite{maggiolino2023deep}            & 26.7 & 24.6 & 29.6 \\
Imprassoc~\cite{stadler2023improved}            & 26.9 & 24.6 & 28.6 \\
OC-SORT~\cite{cao2023observation}               & 23.2 & 21.6 & 26.4 \\
StrongSORT~\cite{du2023strong}                  & 27.8 & 24.4 & 31.7 \\
BoTSORT~\cite{aharon2022bot}                    & \underline{29.6} & \underline{28.0} & 34.3 \\
\rowcolor{gray!15}
\textbf{BoTSORT FT}                     & \textbf{45.7}\,(+\textbf{16.1}) & \textbf{58.8}\,(+\textbf{30.8}) & \textbf{56.1}\,(+\textbf{21.8}) \\
\hline
\end{tabular}

\end{table}
\textbf{Quantitative Results}
Most Multi-Object Tracking models follow Detection-by-Tracking paradigm to enhance their tracking performance.
In this experiment, we benchmark several detection-by-tracking models.
\Cref{tracking_result} shows quantitative results on our dataset.
The values in parentheses indicate the improvement compared to the corresponding pre-trained model. 

Among all models using pretrained components, BoT-SORT achieves the best performance across all three metrics.
Therefore, we selected BoT-SORT as the base tracker and integrated our fine-tuned detection model to evaluate the effect of fine-tuning.
We used the same train/val/test splits as in the detection experiments.

The result show that fine-tuned BoT-SORT significantly outperforms all other methods, achieving 45.7 HOTA, 58.8 MOTA, and 56.1 IDF1.
These scores represent large improvements of +16.1, +30.8, and +21.8 respectively over the vanilla BoT-SORT, demonstrating the effectiveness for downstream tracking tasks.

\subsection{Ablation Study}
We demonstrate that fine-tuning significantly enhances tracking performance.
In this section, we further analyze the effects of fine-tuning, considering that most existing MOT models rely on both detection and re-identification components.

\noindent
\textbf{Detection and ReID fine-tuning effect}:
\Cref{tab:ablation} represents BoTSORT's tracking performance under four configurations: with and without fine-tuning of the detection and re-identification (ReID) models.
Joint fine-tuning of both components leads to the best results, increasing HOTA from 29.6 to 45.7 (+16.1), MOTA from 28.0 to 58.8 (+30.8), and IDF1 from 34.3 to 56.1 (+21.8) compared to the untuned baseline.
It is notable to mention that the majority of the increases stem from detection fine-tuning alone, which boosts HOTA by +15.3 and MOTA +30.2, even without any changes to the ReID model.
In contrast, ReID fine-tuning offers only marginal improvements: +0.3 in HOTA, +0.0 in MOTA, and +0.2 in IDF1.
These results highlight that, in American football scenarios, where frequent occlusions and dense player formation are common, accurate detection plays a more critical role in tracking performance than identity embedding refinement.

\begin{table}
\caption{Tracking performance using BoT-SORT with combinations of fine-tuned (FT) detection and ReID models. We evaluate the effect of fine-tuning each component independently and jointly. Bold = best performance per metric.}
\label{tab:ablation}
\centering
\scriptsize
\setlength{\tabcolsep}{3pt}
\resizebox{\columnwidth}{!}{
\begin{tabular}{c|c|c|c|c}
\hline
\textbf{Detection FT} & \textbf{ReID FT} & \textbf{HOTA↑} & \textbf{MOTA↑} & \textbf{IDF1↑} \\
\hline
  &        & 29.6 & 28.0 & 34.3 \\
  &  \checkmark  & 29.9 (+0.3) & 28.0 (+0.0) & 34.5 (+0.2) \\
 \checkmark & & 44.9 (+15.3) & 58.2 (+30.2) & 55.3 (+21.0) \\
 \checkmark & \checkmark & \textbf{45.7} (+\textbf{16.1}) & \textbf{58.8} (+\textbf{30.8}) & \textbf{56.1} (+\textbf{21.8}) \\
\hline

\end{tabular}
}

\end{table}

\noindent
\textbf{Detection Models}:
For detection, we compared YOLO and RTDETR models.
Their detailed experimental settings are described in Section ~\ref{ex_detection}.

\begin{table}
\caption{Ablation on detection fine-tuning (FT) with tracker and reid model fixed. Parentheses show differences between pre-trained and fine-tuned model. Bold = best performance per metric.}
\label{ablation_detection}
\centering
\scriptsize
\setlength{\tabcolsep}{3pt}
\label{ablation}
\resizebox{\columnwidth}{!}{
\begin{tabular}{c|c|c|c|c}
\hline
\textbf{Model} & \textbf{FT} & \textbf{HOTA↑} & \textbf{MOTA↑} & \textbf{IDF1↑} \\
\hline
\multirow{2}{*}{RTDETR} & & 32.5 & 30.4 & 38.1 \\
                        & \checkmark & 45.0 (+12.5) & 56.9 (+26.5) & 56.9 (+18.8) \\
\hline
\multirow{2}{*}{YOLO}   &  & 29.9 & 28.0 & 34.5 \\
                        & \checkmark & \textbf{45.7 (+15.8)} & \textbf{58.8 (+30.8)} & \textbf{57.9 (+23.4)} \\
\hline
\end{tabular}
}

\end{table}

\Cref{ablation_detection} shows the BoTSORT performance with and without fine-tuned detection model.
We utilized RTDETR-l and YOLOv10m for comparison, as they exhibited the highest performance among YOLO and RTDETR models.

The results (\Cref{ablation_detection}) indicate that fine-tuning significantly improves tracking performance across all metrics for both models.
In particular, YOLO sees a substantial MOTA increase of +30.8, outperforming RTDETR's +26.5 gain.
Although RTDETR models outperform YOLO in detection (see \Cref{quantitative_result}), YOLO achieves superior results in the MOT setting,

\noindent
\textbf{Re-Identification Models}:
Additionally, we fine-tuned OSNet \cite{zhou2019omni} and CLIP \cite{radford2021learning} as part of our ablation study.
OSNet is a CNN-based deep learning network designed for the re-identification task and remains a strong baseline even with the advent of transformer-based models.
For the transformer-based approach, we fine-tuned CLIP, which employs Vision Transformer architecture.
We fine-tuned both OSNet and CLIP on our dataset and then used each as a re-identification feature extractor to assess their impact on tracking performance.

\begin{table}
\caption{Ablation on ReID fine-tuning (FT) with tracker and detection model fixed. Parenthesis shows differences between pre-trained and fine-tuned model. Bold = best performance per metric.}
\label{ablation_reid}
\centering
\begin{tabular}{c|c|c|c|c}
\hline
\textbf{Model} & \textbf{FT} & \textbf{HOTA↑} & \textbf{MOTA↑} & \textbf{IDF1↑} \\
\hline
  \multirow{2}{*}{CLIP} & & 45.6 & 58.2 & \textbf{56.1} \\
   &  \checkmark  & 45.4 (-0.2) & 58.1 (-0.1) & 55.7 (-0.4) \\
   \hline
  \multirow{2}{*}{OSNet} & & 44.9 & 58.2 & 55.3 \\
   & \checkmark  & \textbf{45.7 (+0.8)} & \textbf{58.8 (+0.6)} & \textbf{56.1 (+0.8)}\\
\hline
\end{tabular}

\end{table}

As shown in \Cref{ablation_reid}, fine-tuning OSNet slightly improve HOTA by +0.8, MOTA by +0.6, and IDF1 by +0.8.
In contrast, CLIP showed a slight performance degradation after fine-tuning, with decreases of -0.2 in HOTA, -0.1 in MOTA, and -0.4 in IDF1.
Notably, the lightweight CNN-based OSNet performed better than the more resource-intensive transformer-based CLIP model.
This suggests that model simplicity and domain compatibility may outweigh architectural complexity in sports tracking tasks, especially where appearance features are noisy or occlusion-prone.

\noindent
\textbf{MOT models}:
Lastly, we compare the tracking capabilities of multiple MOT models using the same detection and re-identification backbones.
We use the fine-tuned YOLOv10m model as our detector, and the fine-tuned OSNet model as our re-identification feature extractor model.

\begin{table}
\caption{Tracking performance with fine-tuned detection model and re-identification model, Underline = Second best performance per metric}
\label{ablation_mot}
\centering
\begin{tabular}{l|c|c|c}
\hline
\textbf{Method} & \textbf{HOTA↑} & \textbf{MOTA↑} & \textbf{IDF1↑} \\
\hline
BoostTrack~\cite{stanojevic2024boosttrack}            & \underline{40.7} & 53.3 & 50.9 \\
ByteTrack~\cite{zhang2022bytetrack}             & 40.6 & 55.3 & 49.3 \\
DeepOCSORT~\cite{maggiolino2023deep}            & 41.3 & 57.0 & 51.3 \\
Imprassoc~\cite{stadler2023improved}            & 39.3 & 51.9 & 44.0 \\
OC-SORT~\cite{cao2023observation}               & 37.7  & 55.7 & 46.7  \\
StrongSORT~\cite{du2023strong}                  &  43.6 &  \underline{58.4} & \underline{54.9} \\
BoTSORT~\cite{aharon2022bot}                    & \textbf{45.7}\ & \textbf{58.8}  & \textbf{56.1}\ \\
\hline
\end{tabular}

\end{table}

\Cref{ablation_mot} illustrates BoT-SORT outperforms all other MOT models across all metrics.
This demonstrates its robustness and suitability for the American football scenario, where crowded scenes and occlusion are common.
StrongSORT ranks second in both MOTA (58.4) and IDF (54.9), where the gap is over 2 points (43.6 vs 45.7).
Other models such as DeepOCSORT and ByteTrack show relatively strong results, but lag behind in IDF1, indicating challenges in preserving identities over time.

\section{Conclusion}
We introduced the first American football Multi-Object (MOT) Dataset with human annotation.
We benchmark our dataset using existing detection-by-tracking based MOT algorithms and analyze the impact of fine-tuning both object detectors and re-identification models.
For the detection and tracking, we demonstrated fine-tuned models outperform pre-trained models significantly.
From these results, our dataset and evaluation can serve as the foundation of future American football video analysis and contribute to MOT performance.

\begin{acks}
The authors would like to express their gratitude to the members of the Kantoh Collegiate American Football Association, including the Keio University American Football Team (Unicorns), for providing the game videos.
\end{acks}

\bibliographystyle{ACM-Reference-Format}
\balance
\bibliography{reference}

\end{document}